# Population structure-learned classifier for high-dimension low-sample-size class-imbalanced problem

Author: Liran Shen[1], Meng Joo Er[1], Qingbo Yin[2, *]

[1] College of Marine Electrical Engineering, Dalian Maritime University, Dalian, 116023, China

[2] College of Information Science and Technology, Dalian Maritime University, Dalian, 116023, China Dalian

*Correspondence and requests for materials should be addressed to Qingbo Yin (Email: qingbo@dlmu.edu.cn)

# Abstract

The Classification on high-dimension low-sample-size data (HDLSS) is a challenging problem and it is common to have class-imbalanced data in most application fields. We term this as Imbalanced HDLSS (IHDLSS). Recent theoretical results reveal that the classification criterion and tolerance similarity are crucial to HDLSS, which emphasizes the maximization of within-class variance on the premise of class separability. Based on this idea, a novel linear binary classifier, termed Population Structure-learned Classifier (PSC), is proposed. The proposed PSC can obtain better generalization performance on IHDLSS by maximizing the sum of inter-class scatter matrix and intra-class scatter matrix on the premise of class separability and assigning different intercept values to majority and minority classes. The salient features of the proposed approach are: (1) It works well on IHDLSS; (2) The inverse of high dimensional matrix can be solved in low dimensional space; (3) It is self-adaptive in determining the intercept term for each class; (4) It has the same computational complexity as the SVM. A series of evaluations are conducted on one simulated data set and eight real-world benchmark data sets on IHDLSS on gene analysis. Experimental results demonstrate that the PSC is superior to the state-of-art methods in IHDLSS.

# 1. Introduction

Recent accumulation of high-dimension data sets have accelerated the interest in the development of class prediction or new insights for classification. Typical applications include computer vision, medical image analysis, disease diagnostics and chemometrics, especially gene analysis. However, when the sample size is less than the feature dimension $d$ ($n << d$, high-dimensional low-sample-size: HDLSS), classical statistical methods encounter the performance degradation for classification[1].

The related study on HDLSS can be summarized into two research routes according to whether the dimensions of the features were reduced or not [1, 2]. The first route takes advantage of regularity or dimensionality reduction as specific preprocessing steps. The methods in this route involve the majority of modern classifiers, i.e., Discriminant Analysis [3], Mean Difference (MD) [4], Ensemble Learning [5, 6], Penalized logistic regression (PLR) [7, 8], Neural Networks (NN) [9] and Deep Learning [10, 11]. Although there are many successful applications in specific scenarios, these methods for classification are often subjected to serious drawbacks of having biased discriminant scores due to  (1) the assumption of feature independence, (2) unstable estimation of high-dimension covariance matrices, (3) being infeasible when both of the dimension of data and the sample size are very large [1]. The second route only studies the methods without consideration of any dimensionality reduction, which imply that these methods work straightforwardly on HDLSS data sets for classification. In this route, there are few methods with geometry representation. Maximum Margin Criterion (MMC) [12] is a variant of linear discriminant analysis, which avoids solving the inverse of the low rank between-class scatter matrix for HDLSS. Support Vector Machine (SVM) [13] is a universal classifier, which maximizes the smallest distance between classes and can be used directly to any data set, regardless of whether $n$ is larger or smaller than $d$. These methods lead to a phenomenon of data overfitting, so-called "data-piling". The distance-weighted methods (DWD, wDWD and DWSVM) were proposed to improve the SVM in the HDLSS setting [14-17], which maximize harmonic mean between classes with more computing consumption due to second-order cone programming (SOCP) than quadratic programming for SVM. PGLMC [1] was conceived to combine the local structure of the hyperplane and the global statistics information of population with same computational complexity owing to solving similar Quadratic Programming (QP) formulation as SVM. NPDMD [18] adopted the classification criterion for HDLSS, tolerance similarity, to maximize the intra-class (or within-class) variance on the premise of class separability.

For HDLSS data sets, class-imbalanced data are common in most application fields (especially biomedical field). We denote it as IHDLSS (Imbalanced HDLSS). The standard classifiers assume that it is an equal cost of misclassification in each class, and tend to identify or assign most of the new samples to the majority class and obtain very poor accuracy for the minority class even when the imbalanced factor $m$ is only moderate, which refers to the ratio of the majority class size to the minority class size. This bias from class-imbalanced data becomes an additional challenge for classification on HDLSS data sets [1, 19]. The few studies that focus on IHDLSS mainly leverage the combination of feature selection (dimension reduction) and correction strategies to account for differential class sizes [19, 20]. In this paper, a new cost-sensitive linear binary classifier is proposed to address the class-imbalanced problem on HDLSS without consideration of dimensionality reduction. This method is denoted as Population Structure-learned Classifier (PSC), which pursues to maximize the sum of between-class scatter matrix and within-class scatter matrix on the premise of class separability, and assigns different intercept values to majority and minority classes. This method is implemented on quadratic programming.

The rest of this paper is organized as follows. Section 2 presents the related methods and their characteristics. Section 3 presents on the proposed PSC. Section 4 presents the experimental results and discussions. Finally, Section 5 concludes the paper.

## 2. The drawbacks of the related methods

The severe overfitting phenomenon ("data-piling") will appear in the HDLSS data setting for SVM [17] and some other classifiers [15]. Data-piling means that most of the data is concentrated in two parallel hyperplanes [1, 21]. Marron and Qiao et al. proposed the distance-weighted methods (DWD, wDWD and DWSVM) to improve the SVM in the HDLSS setting [14-17]. These distance-weighted methods maximize harmonic mean between classes by second-order cone programming (SOCP), which demands more computationally consumption than quadratic programming for SVM [22-24]. In Ref [1], PGLMC adopted the first-order statistics information of training data to map the data over as large an interval as possible in the projection space by the similar QP formulation as SVM. Although the above methods alleviate the data-piling, it is still inevitable to suffer from this overfitting issue for HDLSS.

For class-imbalanced problem, most classification algorithms adopted the correction strategies to make up the imbalanced detection rate across two classes [25, 26]. Those strategies fall into two types, namely sampling methods and weighting methods. The first type deals with class-imbalanced data by either over-sampling the minority class or under-sampling the majority class. The second type is referred to as cost-sensitive learning, which improves the detection rate of the minority class by adjusting the weighting or decision threshold in classification on the imbalanced data [27]. But for IHDLSS, there are only few methods (cost-sensitive SVM, distance-weighted methods and PGLMC), all of which come from cost-sensitive learning [1, 28, 29].

For binary classification problems, we denote the linear discriminant function as $f(x) = (w^T x + b)$, which maps a data point $x \in R^d$ to a class label $y \in \{+1, -1\}$, where the direction vector $w \in R^d$ has unit $L_2$ norm, and $b \in R$ is the intercept term.

2.1 SVM

The objective function of the soft-margin SVM is as follows:

$$\underset{w,b,\varepsilon_i}{\operatorname{argmin}} \frac{1}{2} \|w\|^2 + C \sum_{i=1}^{N} \varepsilon_i \tag{1}$$

$$\text{s.t. } y_i(w^T x_i + b) \geq 1 - \varepsilon_i, \ \varepsilon_i \geq 0, i = 1, 2, \cdots, N$$

where $\varepsilon_i$ is the slack variable. The standard definition is for all cases, but with a little unsuitable for class-imbalanced data set. The cost-sensitive SVM (csSVM) is formulated as follows

$$\underset{w,b,\varepsilon_i}{\operatorname{argmin}} \frac{1}{2} \|w\|^2 + C \sum_{i=1}^{N} \tau_i \varepsilon_i \tag{2}$$

$$\text{s.t.} \ y_i(w^T x_i + b) \geq 1 - \varepsilon_i, \ \varepsilon_i \geq 0, i = 1,2,\cdots, N$$

where $\tau_i$ is the cost-sensitive parameter for $i$th sample. It can be found that csSVM attempts to balance the training error between two classes, and improve the performance for class-imbalanced data set to some extent. But, for both of SVM and csSVM, it is true to undergo the data-piling phenomenon and a loss of generalizability in HDLSS and IHDLSS settings [1, 30]. For the detailed proof about data-piling of SVM, please refer to [30]. In the sequel, for class-imbalanced problem, the notation SVM actually refer to csSVM.

2.2 The distance-weighted methods

The original objective function of the Distance-weighted discrimination (DWD) is

$$\underset{w,b}{\text{argmin}} \sum_{i=1}^{n} \left(\frac{1}{r_i} + C\eta_i\right) \tag{3}$$

$$\text{s.t.} \ r_i = y_i(x_i^T w + b) + \eta_i, \ r_i \geq 0, \eta_i \geq 0, \quad \|w\|^2 \leq 1$$

The DWD method is sensitive to the imbalanced data [16]. To circumvent the constraint of imbalanced data, the weighted DWD (wDWD) [16] was proposed as follows

$$\underset{w,b}{\text{argmin}} \sum_{i=1}^{N} \mathcal{W}(y_i) \frac{1}{r_i} \tag{4}$$

$$\text{s.t.} \ \|w\|^2 \leq 1, r_i \geq 0, \ r_i = y_i(x_i^T w + b), i = 1,2,\cdots, N.$$

where $\mathcal{W}(y_i)$ is the weight or cost-sensitive parameter for the $i$th training sample. Although weighted DWD has improved standard DWD for imbalanced data and various nonstandard situations, wDWD is still with heavy computing consumption due to SOCP. Recent theoretical results show that it is still inevitable for distance-weighted methods to suffer from data-piling [18].

2.3 PGLMC

The objective function of the PGLMC[1] is as follow:

$$\underset{w}{\operatorname{argmin}} \left( \frac{\|w\|^2}{(m_1-m_2)^T w} + C_0 \sum_{i=1}^{N} \xi_i \right) \tag{5}$$

$$\text{s.t. } y_i(w^T x_i + b) \geq 1 - \xi_i, i = 1,2,\cdots,N$$

where $m_i$ is the mean of training samples from $i$th class, $i = 1,2$. The PGLMC combines the local structure of the hyperplane and the first-order statistics information of population to construct more stable margin between two classes than SVM. The PGLMC merely uses the item $(m_1 - m_2)$ to control the differences between two classes, and does not consider the intra-class differences. Therefore, as the methods based on Distance Weighting do, the PGLMC only alleviates the issue of data-piling instead of overcoming it.

2.4 NPDMD

The NPDMD is conceived to emphasize maximization of within-class variance on the premise of class separability [18]. The objective function of the NPDMD is as follow:

$$\underset{w}{\min} \left( \frac{\|w\|^2}{w^T S_W w} + C_0 \sum_{i=1}^{n} \xi_i \right) \tag{6}$$

$$\text{s.t. } y_i(w^T x_i + b) \geq 1 - \xi_i, i = 1,2,\cdots,n$$

$$S_W = \Sigma_1 + \Sigma_2, \quad \Sigma_j = \frac{1}{n_j} \sum_{x \in j_{class}} (x - u_j)(x - u_j)^T$$

where $u_j$ is the mean of training samples from $j$th class. The NPDMD avoids data-piling and exhibits superior performance in HDLSS. However, the class-imbalanced problem is out of its insight.

3. Population Structure-learned Classifier

3.1 Motivation

For IHDLSS, there are two issues, namely data-piling and class-imbalanced problems (which seriously degrade the classification performance), that must be addressed. As foreshadowed, most of the aforementioned methods only leverage local information of sample population. For example, SVM only uses the support vectors to construct the projection direction. For HDLSS, the boundary of two classes derived from SVM are unstable [1, 31]. The cause of data-piling is that the methods merely mention the inter-class difference and the intra-class similarity $min_w(w^T S_w w)$, but neglect the intra-class difference[18]. In practical applications, when the sample size is enough ($d \ll n$), these methods have fine performance. In the case of HDLSS, they are biased and not stable because there are plenty of disturbing clues to meet the intra-class similarity criteria $min_w(w^T S_w w)$, which only emphasizes the maximization of similarity. In this paper, the classification criterion for HDLSS, tolerance similarity of *[18]*, is adopted, which includes two rules: (1) Separable. In theory, assume that there is at least a hyperplane to separate clearly the samples to two classes. (2) The similarity and difference of intra-class samples. In view of this, on the premise of class separability, $max_w(w^T S_w w)$ is a good choice on HDLSS instead of $min_w(w^T S_w w)$ to measure the similarity with tolerance difference.

For the class-imbalanced problem, a two-dimension illustration is used to express the characteristic of the above methods in Fig. 1. The data are generated from a multivariate normal distribution $N_d(\pm\mu, \Sigma)$ for two classes, where $d = 2$, $\mu = (1, 2.5)$ and $\Sigma = [1.5\ 0.5; 0.5\ 1.5]$. In Fig.1(a), there are 5 positive samples (blue) and 65 negative (red) samples, $m = 13$. In Fig.1(b), there are 12 positive samples and 65 negative samples, $m = 5.42$. In Fig.1(c), there are 32 positive samples and 65 negative samples, $m = 2.03$. In Fig.1(d), there are 65 positive samples and 65 negative samples, $m = 1$. The blue solid dot represents the mean vector of positive samples, and the red solid dot for the mean vector of negative samples. The blue ellipse represents the possible scope of positive samples with 95% confidence. The red ellipse indicates the possible scope of negative samples with 95% confidence. In this example, the Bayes rule has the intercept $b_{Bayes} = 0$ and the direction $w_{Bayes} = \Sigma^{-1}(2\mu)$. The Bayes rule [32] serves as the benchmark for comparison in theory, and is formulated as $w_{Bayes} = \Sigma^{-1}(\mu_+ - \mu_-)$, $b_{Bayes} = -1/2 \, w_{Bayes}^T (\mu_+ + \mu_-)$. Here, positive population mean $\mu_+ = \mu$ and negative population mean $\mu_- = -\mu$. The real mean difference direction (RMDD) is obtained by the vector $(u_1 - u_2)$, instead of $(\mu_+ - \mu_-)$ in the Bayes rule. However, because the theory distribution of the sample population cannot be known in real-world applications, RMDD is a more valuable benchmark than $w_{Bayes}$. The decision boundary of SVM is the cyan dashed line. The decision boundary of PGLMC is the magenta dash-dot line. The decision boundary of the NPDMD is the red point solid line. As shown in Fig.1, while we increase the positive samples from 5 to 65, it can be found that (1) The real mean difference direction is robust; (2) The decision boundary of SVM suffers from a series of apparent changes, and cannot converge to the mean difference direction; (3) The decision direction of PGLMC is also robust with a little different intercept; (4) The decision direction of the NPDMD gradually converges to the mean difference direction with imbalanced factor $m$ from 13 to 1. In view of the observations in Fig. 1, two operations are adopted to correct the bias caused by the class-imbalanced problem, which involves (1) the mean difference vector $(u_1 - u_2)$ of training data is a robust reference for projection direction, and may be combined with $w^T S_w w$ to formulate a more reasonable object function; (2) The cost-sensitive process is a valuable compensation for intercept $b$. In Fig. 1(d), it is shown that our method

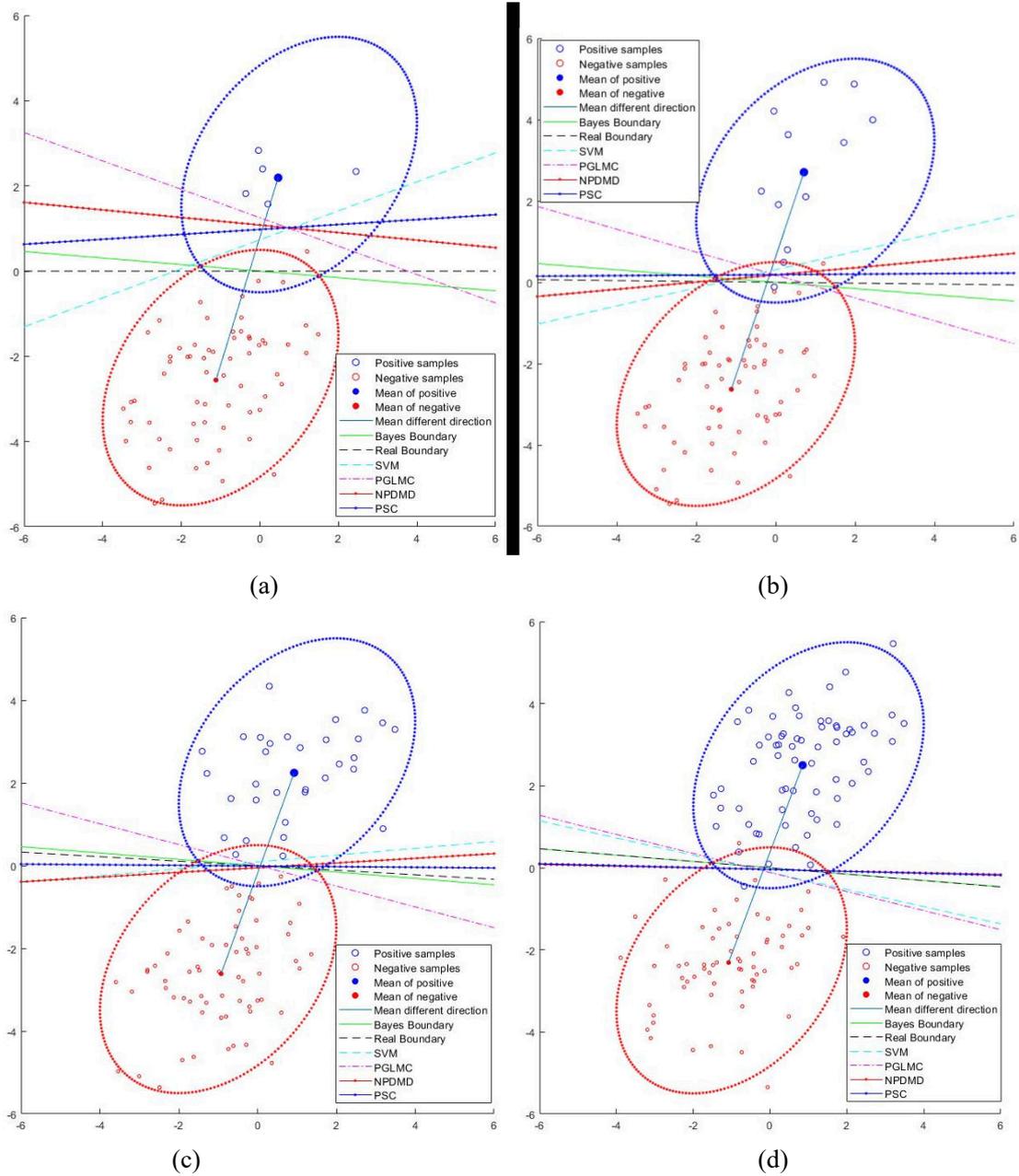

Figure 1. The illustration of border variability for two classes on class-imbalanced data set. (a)5 positive samples and 65 negative samples. (b) 12 positive samples and 65 negative samples. (c) 32 positive samples and 65 negative samples. (d) 65 positive samples and 65 negative samples.

converges to RMDD fastest among four methods.

3.2 Population Structure-learned Classifier

Inspired by the above discussions, for IHDLSS, a linear classifier is proposed to combine the mean difference vector and the intra-class scatter matrix to construct a new object function, which is denoted as Population Structure-learned Classifier (PSC).

$$\min_w \left( \frac{\|w\|^2}{\beta w^T S_B w + w^T S_W w} + C_0 \sum_{i=1}^n \mathcal{W}(y_i) \xi_i \right) \quad (7)$$

$$\text{s.t. } y_i(w^T x_i + b) \geq 1 - \xi_i, i = 1, 2, \cdots, n$$

where

$$\xi_i = \ell(y_i f(x_i)), \xi_i \geq 0, \mathcal{W}(y_i) = \begin{cases} 1, & y_i == 1 \\ \frac{n_1}{n_2}, & y_i == -1 \end{cases}$$

$$S_W = \frac{1}{n_1} \sum_{x \in class\ 1} (x - u_1)(x - u_1)^T + \frac{1}{n_2} \sum_{x \in class\ 2} (x - u_2)(x - u_2)^T \quad (8)$$

$$S_B = (u_1 - u_2)(u_1 - u_2)^T \quad (9)$$

$$\beta = \begin{cases} e^{\left[-\log\left(\frac{n_1}{n_2}\right)/4\right]}, & \frac{n_1}{n_2} \geq 1 \\ e^{\left[-\log\left(\frac{n_2}{n_1}\right)/4\right]}, & \frac{n_2}{n_1} > 1 \end{cases} \quad (10)$$

where $u_j = (1/n_j) \sum_{x \in class\ j}(x)$, $n_j$ is the sample size for class $j$, $j = 1,2$. And $\ell(\vartheta)$ is the hinge loss $\ell(\vartheta) = max(0, 1 - \vartheta)$. The item $\|w\|^2$ in the numerator is minimized to separate the samples from two classes. The term $\beta w^T S_B w + w^T S_W w$ in the denominator is maximized to ensure that the samples from two classes are not only separated, but also far away from each other in the projecting space.

It has been proven that $1/(\beta w^T S_B w + w^T S_W w)$ and $[C - w^T(\beta S_B + S_W)w]$ are with same effect in the optimization formula [2]. Eq. (7) can be reformulated to facilitate calculation as follows:

$$\min_{w} \left\{ \frac{1}{2}\|w\|^2 + \frac{1}{2}\lambda[C - w^T(\beta S_B + S_W)w] + \alpha^T[\mathbf{1} - \xi - Y(Xw + b\mathbf{1})] + C_0\xi^T\mathcal{W} \right.$$
$$\left. - \mu^T\xi \right\}$$

(11)

where $Y$ is the $N \times N$ diagonal matrix with the components of $y_i$ on its diagonal; $X \in R^{N \times d}$, which $i$th row is sample $x_i$; $w \in R^{d \times 1}$ is the direction vector or projecting vector; $\mathbf{1}$ is the column vector 1; $\alpha = (\alpha_1, \cdots, \alpha_N)^T \in R^{N \times 1}$, $\alpha_i > 0$ and $\alpha_i$s are Lagrangian multipliers; $\mu = (\mu_1, \cdots, \mu_N)^T \in R^{N \times 1}$, $\mu_i > 0$ and $\mu_i$s are Lagrangian multipliers; $\xi = (\xi_1, \cdots, \xi_N)^T \in R^{N \times 1}$.

By differentiating the Lagrangian formulation with respect to $w$, $b$ and $\xi$, we obtain the following conditions:

$$\frac{\partial L}{\partial w} = w - \lambda(\beta S_B + S_W)w - X^T Y^T \alpha = 0$$
$$w = [I - \lambda(\beta S_B + S_W)]^{-1} X^T Y^T \alpha \tag{12}$$

$$\frac{\partial L}{\partial \xi} = C_0 \mathcal{W} - \alpha - \mu, \quad C_0 \mathcal{W} = \alpha + \mu \tag{13}$$

$$\frac{\partial L}{\partial b} = \alpha^T Y \mathbf{1} = 0 \tag{14}$$

When substituting (12), (13) and (14) into (11), we can obtain the dual formulation as follows

$$L(\alpha) = -\frac{1}{2}\alpha^T YX[I - \lambda(\beta S_B + S_W)]^{-1} X^T Y^T \alpha + \alpha^T \mathbf{1} + \frac{1}{2}\lambda C \tag{15}$$

Given that

$$G = YX[I - \lambda(\beta S_B + S_W)]^{-1} X^T Y^T \tag{16}$$

where $G$ is a symmetric positive semidefinite matrix, eq. (15) becomes

$$L(\alpha) = -\frac{1}{2}\alpha^T G \alpha + \alpha^T \mathbf{1} \tag{17}$$

Hence, the optimization problem (7) can be reformulated to the following

$$\underset{\alpha}{\mathrm{argmax}}\, L(\alpha) \tag{18}$$

$$\text{s.t.}\ \alpha^T Y \mathbf{1} = 0, C_0 \mathcal{W} \geq \alpha_i \geq 0$$

The above formula is a classical quadratic programming problem. The Karush-Kuhn-Tucker (KKT) condition should be as follows:

$$C_0 \mathcal{W} \geq \alpha_i \geq 0,\ \mu_i \geq 0$$

$$y_i(w^T x_i + b) - 1 + \xi_i \geq 0$$

$$\alpha_i [y_i(w^T x_i + b) - 1 + \xi_i] = 0$$

$$\xi_i \geq 0,\ \mu_i \xi_i = 0$$

Eq. (18) can be regarded as a quadratic programming problem with equality constrains while inequality conditions are just looked upon as to scale the coefficients $\alpha_i$.

3.3 The estimated intercept

Note that for the linear discriminant function, there are two parameters $(w, b)$ to estimate. The intercept $b$ is with the same importance as the discriminant direction $w$ for the classification/prediction performance [32]. For IHDLSS setting, the boundary space of minority (even majority) class is unstable and underestimated. Therefore, the intercept $b$ from this underestimated boundary is biased and results in poor performance. The following Lemma and Theorem give the relationship of the boundary for two classes.

**Theorem 1**. Assume that the data are sampled from the population with probability density function of Gaussian $N(\mu, 1)$. For two independent sampling progresses, the data distribution interval of the sampling process with more sampling points is larger than that with less sampling points under the same confidence.

**Proof**:

For any given confidence $\rho$, $(0 < \rho < 1)$, there is an arbitrary positive $\exists \varepsilon \in (0, \infty)$ and the following formulation is true

$$p(|x - \mu| < \varepsilon) > 1 - \rho \qquad (19)$$

About negative class from first sampling, there are $n_-$ data points

$$p(|x_- - \mu| < \varepsilon_-) > 1 - \rho$$

$$[p(|x_- - \mu| < \varepsilon_-)]^{n_-} > (1 - \rho)^{n_-}$$

About positive class from second sampling, there are $n_+ = n_-/m$ data points, $m > 1$.

$$[p(|x_+ - \mu| < \varepsilon_+)]^{n_+} > (1 - \rho)^{n_-}$$

$$[p(|x_+ - \mu| < \varepsilon_+)]^{\frac{n_-}{m}} > (1 - \rho)^{n_-}$$

$$[p(|x_+ - \mu| < \varepsilon_+)] > (1 - \rho)^m$$

$$[p(|x_+ - \mu| < \varepsilon_+)]^{\frac{1}{m}} > (1 - \rho)$$

$$\therefore p(|x_- - \mu| < \varepsilon_-) > [p(|x_+ - \mu| < \varepsilon_+)]$$

$$\therefore \varepsilon_- > \varepsilon_+ \qquad (20)$$

The proof is completed.

Remark: In real applications, we relax the above inequality as $\varepsilon_- \geq \varepsilon_+$.

**Corollary 1**. Providing the data for two class is subject to independent and identical distribution with different parameters (such as different means for Gaussian), the dispersion interval of samples for majority class is more reliable than that for minority class.

The proof is generalized from Theorem 1.

For the intercept $b$, there are two situations to consider.

(1) If $\sum_{i=1}^{n} \xi_i > 0$, it is assumed that the data in the training set can express the boundary for two classes to some extent. Therefore, the intercept term $b$ can be obtained by the Criterion of Minimum misclassified samples as follows:

$$\underset{b}{\mathrm{argmin}}\, J(b) = \sum_{i}^{N} sgn[-1 * y_i(w^T x_i + b)] \tag{21}$$

$$sgn(x) = \begin{cases} +1, & x \geq 0 \\ -1, & x < 0 \end{cases} \tag{22}$$

(2) If $\sum_{i=1}^{n} \xi_i = 0$, there is an obvious gap between two classes in the direction $w$. Then, inspired by Corollary 1, the gap $b_{gap}$ can be represented as

$$b_{gap} = min(w^T x_+) - max(w^T x_-) = b_- + b_+ \tag{23}$$

where $b_-$ is the distance from $max(w^T x_-)$ to the hyperplane, and $b_+$ is the distance from $min(w^T x_+)$ to the hyperplane. If $n_- \geq n_+$, the bigger the $n_-$, the more stable the boundary of negative class. Therefore, $b_-$ should be larger than $b_+$. For the same reason, if $n_+ > n_-$, the bigger the $n_+$, the more stable the boundary of positive class. It should be that $b_-$ is less than $b_+$. The relationship between $b_-$ and $b_+$ is approximately given by

$$\frac{b_-}{b_+} = \begin{cases} e^{\frac{-[log(n_-/n_+)]}{2R}}, & n_- \geq n_+ \\ e^{\frac{[log(n_+/n_-)]}{2R}}, & n_- < n_+ \end{cases} \tag{24}$$

where $R$ is the scale parameter for trade-off, and it is enough to set $R = 2$. By substituting (24) into (23), we obtain

$$b_+ = \begin{cases} \dfrac{b_{gap}}{1+e^{\frac{-[log(n_-/n_+)]}{2R}}}, & n_- \geq n_+ \\ b_{gap} - \dfrac{b_{gap}}{1+e^{\frac{-[log(n_+/n_-)]}{2R}}}, & n_- < n_+ \end{cases} \tag{25}$$

$$b = b_+ - min(w^T x_+) \tag{26}$$

For a new data $x$, the classifier of PSC is expressed as follows:

$$\hat{y} = (w^T x + b) = \{[I - \lambda(\beta S_B + S_W)]^{-1} X^T Y^T \alpha\}^T x + b \tag{27}$$

3.4 Accelerated version

For formulation (18), we must calculate the inverse of $d \times d$ matrix $[I - \lambda(\beta S_B + S_W)]$, which is a time consuming problem for HDLSS ($d \gg n$). So, we must deformalize or refine the involved computation. $(\beta S_B + S_W)$ can be decomposed as follow:

$$\beta S_B + S_W = \beta(u_1 - u_2)(u_1 - u_2)^T + Q^T K Q \tag{28}$$

$$K = diag(\left[\frac{1}{1*n_1}, \cdots, \frac{i}{i*n_1}, \cdots, \frac{n_1}{n_1*n_1}, \frac{n_1+1}{(n_1+1)*n_2}, \cdots, \frac{n_1+n_2}{(n_1+n_2)*n_2}\right])$$

Where $Q$ is a $n \times d$ matrix, which $i$th row is the sample $x_i - u_j$. Then, it can be known

$$\beta S_B + S_W = D^T L_\tau D \tag{29}$$

$$D = \begin{pmatrix} Q \\ (u_1 - u_2)^T \end{pmatrix}, \quad L_\tau = diag([K, \beta])$$

$$[I - \lambda(\beta S_B + S_W)] = I - \lambda D^T L_\tau D \tag{30}$$

When we resort to the Sherman-Morrison-Woodbury (SMW) identity [33]

$$(A + UCV)^{-1} = A^{-1} - A^{-1}U(C^{-1} + VA^{-1}U)^{-1}VA^{-1} \tag{31}$$

to compute the inverse of $[I - \lambda(\beta S_B + S_W)]$, the following formula is used to transform the original matrix to SMW form.

$$[I - \lambda(\beta S_B + S_W)]^{-1} = [I + D^T(-\lambda L_\tau)D]^{-1} \tag{32}$$

Providing $A = I$, $U = D^T$, $C = -\lambda L_\tau$ and $V = D$

$$[I - \lambda(\beta S_B + S_W)]^{-1} = [I + D^T(-\lambda L_\tau)D]^{-1} = I - I(D^T)[(-\lambda L_\tau)^{-1} + DD^T]^{-1}D$$

$$[I - \lambda(\beta S_B + S_W)]^{-1} = I - D^T[(-\lambda L_\tau)^{-1} + DD^T]^{-1}D \qquad (33)$$

From Equation (33), it can be noticed that both of $L_\tau$ and $DD^T$ are the $(n+1) \times (n+1)$ matrix. Furthermore, $d \times d$ matrix $[I - \lambda(\beta S_B + S_W)]^{-1}$ can be calculated by the inverse of $(n+1) \times (n+1)$ matrix $[(-\lambda L)^{-1} + DD^T]^{-1}$. For HDLSS ($d \gg n$), the computation cost of $[I - \lambda(\beta S_B + S_W)]^{-1}$ can be notably reduced.

3.5 Computation Complexity

The computation complexity for the objective function of PSC involve two parts (1) $d \times d$ matrix $[I - \lambda(\beta S_B + S_W)]^{-1}$. (2) Quadratic Programming formulation for Equation (18).

For $d \times d$ positive semidefinite matrix, the $d$ pairs of eigenvalues and eigenvectors can be computed in $O(d * d)$ time [34], [35]. In the HDLSS case, with $d \gg n$, we adopt the accelerated extension of the PSC with the computation cost in $O(n^2 + 2n)$ time for $d \times d$ matrix $[I - \lambda(\beta S_B + S_W)]^{-1}$. QP's running time is $O(n^{1/2})$ iterations, each iteration requiring $O(n^3)$ arithmetic operations on integers [36]. The computation complexity of PSC is with $O_{PSC} \cong O(n^{7/2})$ due to $O_{PSC} = O(n^2 + 2n) + O(n^{1/2})O(n^3)$, which is the same order as that of those methods based on QP, such as SVM, PGLMC and NPDMD.

The SOCP requires $O(n^{1/2})$ iterations, each requiring $O(n^2 max\{n,d\})$ operations with primal-dual interior point method [15, 37]. In the HDLSS case, with $d \gg n$, the computation consumption of the distance-weighted methods would be $O(n^{5/2} max(n,d))$, which is greater than that of those methods based on QP.

## 4. Experiments

In this section, we conduct the experiments on one simulation data and eight real-world classification problems to evaluate the proposed PSC and compare PSC with DWD, wDWD, SVM, PGLMC and NPDMD.

The codes of this paper were programed in MATLAB and R, and runed in Inter I7-9700 Processor 3.6G Hz system with 64GB RAM. For the methods based on Distance Weighting, we adopt the linear binary implementation in R package 'kerndwd'[20].

### 4.1 Measures of Performance

To evaluate and compare the performance of different methods, we employed some performance measure, such as confusion matrix, ROC curve, correct classification rate (CCR), mean within-group error (MWE) for HDLSS, which were used in [1, 2, 32]. In addition, we design a new measure *balanced correct classification rate* (BCCR) for IHDLSS, as follows:

$$BCCR = \frac{CCR_1 + CCR_2}{2} \cdot e^{\frac{-(CCR_1 - CCR_2)^2}{2}} \qquad (34)$$

where $CCR_1$ is the correct classification rate for class 1 and $CCR_2$ is the correct classification rate for class 2. CCR is a fair measure for the balanced data. Although both of $MWE$ and $BCCR$ are the measures for the imbalanced data, $BCCR$ is a more reasonable measure than $MWE$. The relationship between $BCCR$ and $MWE$ is as follows

$$MWE = \frac{1 - CCR_1}{2} + \frac{1 - CCR_2}{2} = 1 - \frac{CCR_1 + CCR_2}{2} \qquad (35)$$

$$BCCR = \frac{CCR_1 + CCR_2}{2} \cdot e^{\frac{-(CCR_1 - CCR_2)^2}{2}} \qquad (36)$$

where $CCR_1$ is the correct classification rate for the class 1; $CCR_2$ is the correct classification rate for the class 2. $BCCR$ is a more reasonable measure than $MWE$. The relationship between $BCCR$ and $MWE$ is as follows

$$MWE = \frac{1-CCR_1}{2} + \frac{1-CCR_2}{2} = 1 - \frac{CCR_1+CCR_2}{2} \tag{37}$$

$$BCCR = \left[1 - \frac{1-CCR_1}{2} - \frac{1-CCR_2}{2}\right] e^{\frac{-(CCR_1-CCR_2)^2}{2}}$$

$$BCCR = (1 - MWE) e^{\frac{-(CCR_1-CCR_2)^2}{2}} \tag{38}$$

From the above formulation, it can be found that $BCCR$ not only considers the $CCR$s for each class, but also considers the difference between $CCR$s of the two classes.

4.2 Simulations Data: Experiment 1

Providing that specimens from two classes are sampled from multivariate normal distributions $N_d(\pm\mu, \Sigma)$. $\mu \equiv c\mathbf{1}_d$, and $\Sigma \equiv I_d$, where $c > 0$ is a scaling factor with $2c\|\mathbf{1}_d\|_2 = 2.7$. This setting is consistent with the literature [1, 2], which presents a rational difficulty for classification due to the Mahalanobis distance between two classes. For training, there are 100 positive samples and 10 negative samples, (i.e., imbalance factor $m = 10$ ). The sample dimension $d$ varies in {50,80,180,450,800,1600,3200}, thus last five cases definitely correspond to HDLSS setting. The process is in accordance with the literature [32], [17].

Figure 2 shows the experimental results of 15 replications due to six methods on a test dataset with 1500 samples in each class. Figure 2 (a) and (b) are the boxplots of CCRs and the mean curve of CCRs for six methods. The boxplots of DWD express the worst classification performance. When the dimension is 50, the wDWD gets the best CCR, and the PSC follows it. After that, while the dimensionality increases, the CCRs for these six methods gradually tend to be consistent. But, with other dimensions, the PSC gets the best CCRs. Figure 2(c) and (d) are about the boxplots of MWEs and the mean curve of MWEs for six methods. It is obvious that the PSC is the best one among these six methods in the light of MWE. Figures 2(e) and (f) are about the boxplots of BCCRs and the mean curve of BCCRs. Besides the dimension 50, the PSC gets the best performance in most dimensions. In Fig. 2, it can be found that the PSC gradually obtains the performance superiority on all of CCR, WME and BCCR as the dimension increases.

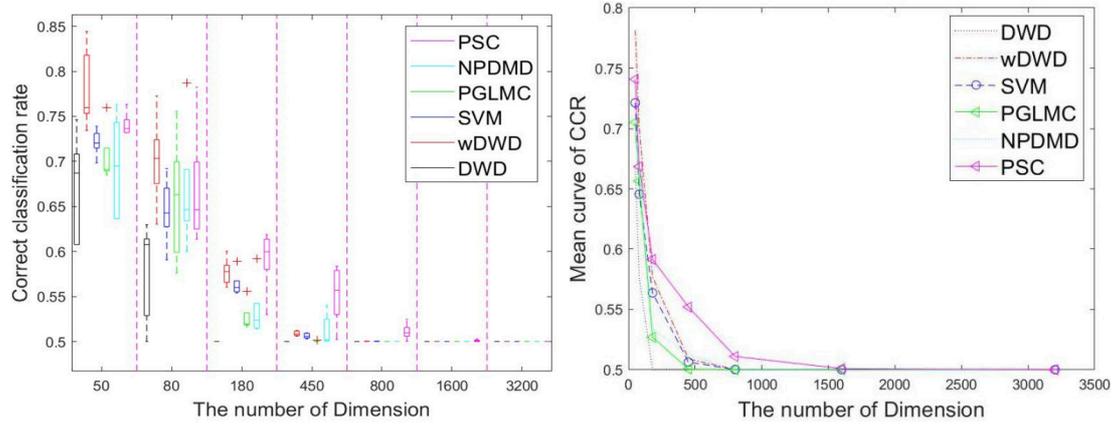

(a)                                   (b)

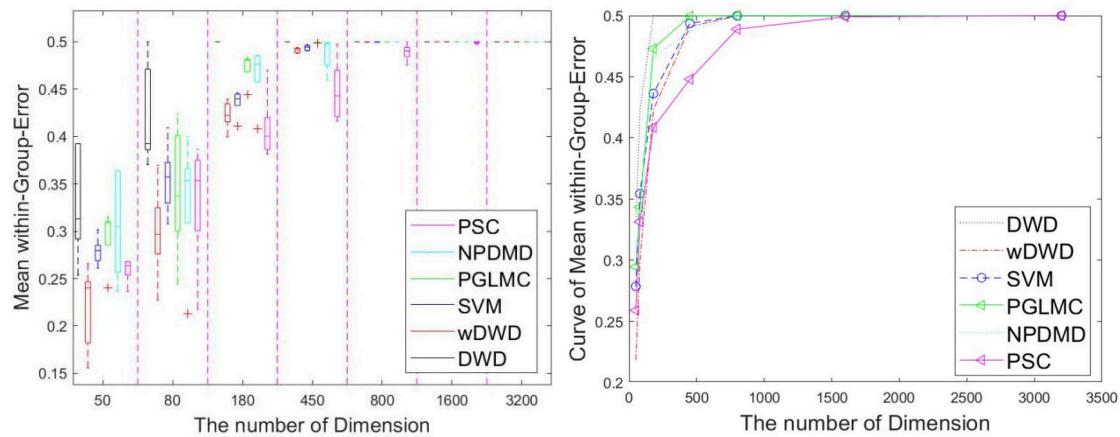

(c)                                   (d)

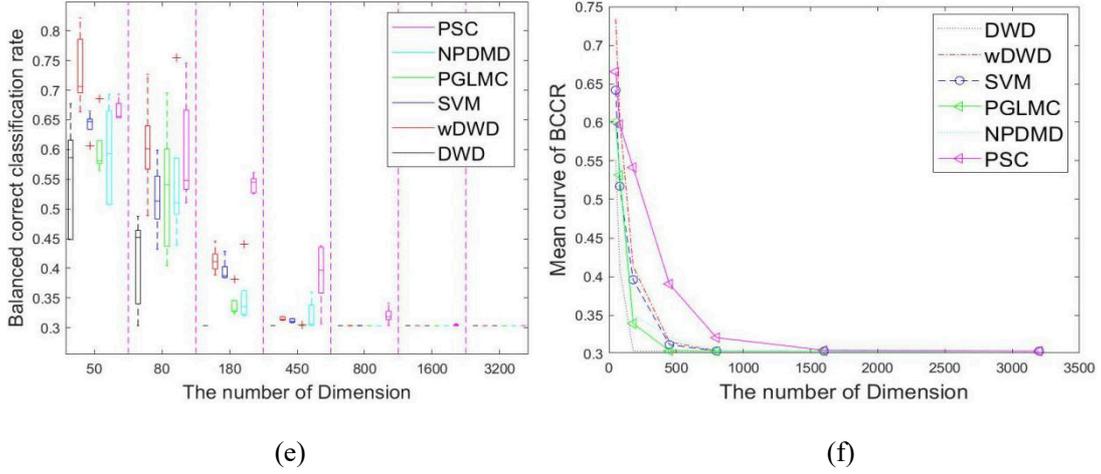

Figure 2. Comparison among six methods for simulation experiment 1 with 5 replications. (a) The boxplots of CCRs. (b) The mean curve of CCRs. (c) The boxplots of MWEs. (d) The mean curve of MWEs. (e) The boxplots of BCCRs. (f) The mean curves of BCCRs.

4.3 Real applications

In the following subsection, the performances of the PSC are evaluated and compared with other five competing classifiers on eight real data sets. The details of eight real data sets are elaborated in Table 1. The data dimensions of these data sets. It is obvious that all of these data sets are for IHDLSS. For the data sets with more classes than 2, we set one class as positive and the rest of the classes as negative to construct an IHDLSS data set with admissible sample size for both classes in the experiments for binary classification.

For all of these data sets, to avoid the randomness of the experimental results, we adopt cross-validation for several times with different randomly sample splits, and report the averages of performance measures. For each data set, all specimens are splitted to five folds, in which 4-folds are for training and one-fold for testing. Parameters for each method are tuned via 4-fold cross-validation within the training data. This process is repeated 18 times.

Table 1. Characteristics of the Data Sets Used in the Experiments

| Data Set | Dim | Class | Positive | Negative | m | Comments |
|---|---|---|---|---|---|---|
| Alon | 2000 | 2 | 22 | 40 | 1.82 | |
| shipp | 7129 | 2 | 58 | 19 | 3.05 | |
| Golub | 7129 | 2 | 47 | 25 | 1.88 | |
| Gordon | 12533 | 2 | 150 | 31 | 4.84 | |
| tian | 12625 | 2 | 137 | 36 | 3.81 | |
| yeoh | 12625 | 6 | 27 | 221 | 8.19 | pos=2; neg=else |
| Burczynski | 22283 | 3 | 101 | 26 | 3.88 | pos=1,2; neg=3 |
| nakayama | 22283 | 10 | 21 | 84 | 4 | pos=5;neg=else |

4.3.1 Experiment 2: Alon data set

In Alon data set [38], there are 40 tumors and 22 normal colon tissues. For each specimen, 6500 genes were expressed with an Affymetrix oligonucleotide array. As the literature [38] does, 2000 genes are retained to be with highest minimal intensity for each specimen.

There are 40 and 22 specimens for 2 classes with imbalance factor $m \approx 1.82$. The performance measures of CCRs, MWEs, BCCR and ROC curve are exhibited in Fig. 3. As shown in Fig.3, the PSC obtains the best AUC, MWE and BCCR. The PGLMC obtains the highest CCR. The confusion matrix is shown in Table 2.

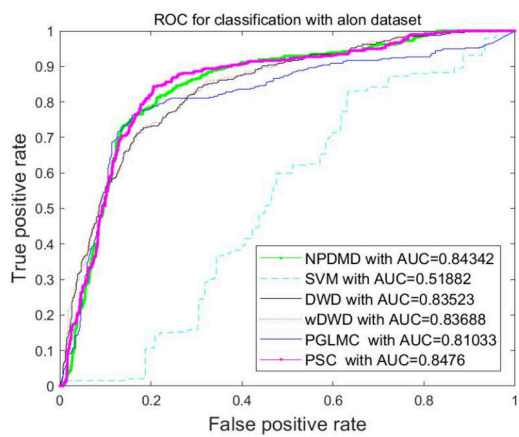
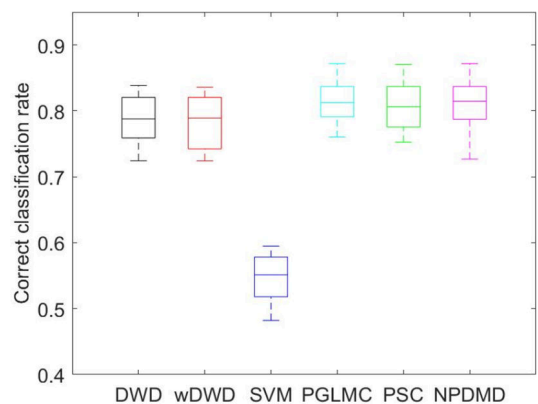

(a)      (b)

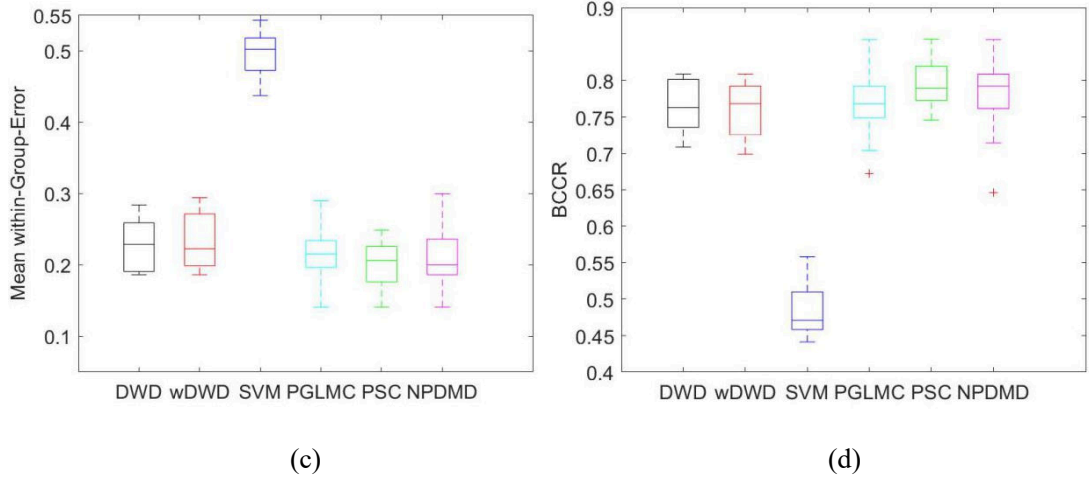

|   | (c) | (d) |
|---|---|---|

Figure 3. Comparison between six methods on Alon data for example 2. (a) ROC curves and AUC. (b) The boxplots of CCRs. (c) The boxplot for the mean within-group error. (d) The boxplot for BCCR.

Table 2. The confusion matrix on Alon data set

| Method | Background | Classification | | CCR | Total CCR | 1-MWE | BCCR |
|---|---|---|---|---|---|---|---|
|  |  | True | False |  |  |  |  |
| DWD | TRUE | 598 | 122 | 0.830556 | 0.78853 | 0.771338 | 0.765948 |
|  | FALSE | 114 | 282 | 0.712121 |  |  |  |
| wDWD | TRUE | 593 | 127 | 0.823611 | 0.78405 | 0.767866 | 0.763109 |
|  | FALSE | 114 | 282 | 0.712121 |  |  |  |
| SVM | TRUE | 464 | 256 | 0.644444 | 0.544803 | 0.50404 | 0.484554 |
|  | FALSE | 252 | 144 | 0.363636 |  |  |  |
| PGLMC | TRUE | 635 | 85 | 0.881944 | **0.812724** | 0.784407 | 0.769623 |
|  | FALSE | 124 | 272 | 0.686869 |  |  |  |
| NPDMD | TRUE | 624 | 96 | 0.866667 | 0.811828 | 0.789394 | 0.780023 |
|  | FALSE | 114 | 282 | 0.712121 |  |  |  |
| PSC | TRUE | 597 | 123 | 0.829167 | 0.807348 | **0.798422** | **0.796914** |
|  | FALSE | 92 | 304 | 0.767677 |  |  |  |

4.3.2 Experiment 3: Shipp data set

In Shipp data set [39], there are 58 diffuse large B-cell lymphomas (DLBCLs) patient specimens, which include 32 positive and 26 negative samples with imbalance factor $m \approx 3.05$. For each specimen, 6,817 gene are expressed with customized cDNA ('lymphochip') microarrays.

The CCRs, ROC and MWEs are shown in Fig. 4. As presented in Fig. 4, DWD and wDWD obtain the best CCR, MWE and BCCR. The PSC gets the best AUC, and the suboptimal of CCR, MWE and BCCR. The corresponding confusion matrix for Shipp data set is exhibited in Table 3.

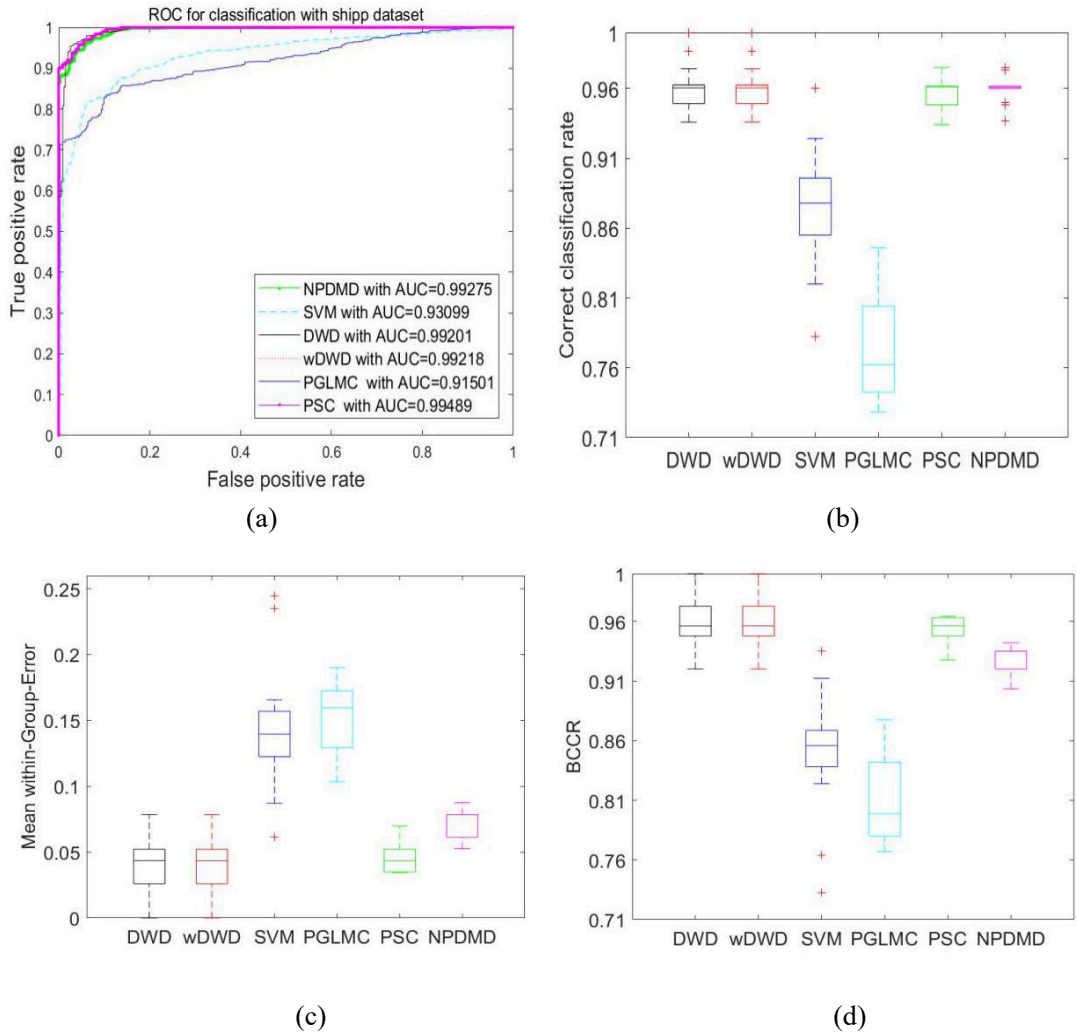

Figure 4. Comparison between six methods on Shipp data for experiment 3. (a) ROC curves and AUC. (b) The boxplots of CCRs. (c) The boxplot for the mean within-group error. (d) The boxplot for BCCR.

Table 3. The confusion matrix on Shipp data set

| Method | Background | Classification | | CCR% | Total CCR | 1-MWE | BCCR |
|---|---|---|---|---|---|---|---|
| | | True | False | | | | |
| DWD | TRUE | 325 | 17 | 0.950292 | **0.962482** | **0.958384** | **0.958258** |
| | FALSE | 35 | 1009 | 0.966475 | | | |
| wDWD | TRUE | 325 | 17 | 0.950292 | **0.962482** | **0.958384** | **0.958258** |
| | FALSE | 35 | 1009 | 0.966475 | | | |
| SVM | TRUE | 280 | 62 | 0.818713 | 0.878788 | 0.85859 | 0.855864 |
| | FALSE | 106 | 938 | 0.898467 | | | |
| PGLMC | TRUE | 341 | 1 | 0.997076 | 0.775613 | 0.850071 | 0.814112 |
| | FALSE | 310 | 734 | 0.703065 | | | |
| NPDMD | TRUE | 302 | 40 | 0.883041 | 0.961039 | 0.934815 | 0.929817 |
| | FALSE | 14 | 1030 | 0.98659 | | | |
| PSC | TRUE | 324 | 18 | 0.947368 | 0.95671 | 0.953569 | 0.953496 |
| | FALSE | 42 | 1002 | 0.95977 | | | |

4.3.3 Experiment 4: Golub data set

In Golub data set [40], there are 38 leukemia patients, which involve 27 specimens for the acute lymphoblastic leukemia and 11 specimens for acute myeloid leukemia with imbalance factor $m \approx 2.45$. For each specimen, 7129 genes are expressed with Affymetrix Hgu6800 chips[41].

The CCRs, MWEs, ROC and BCCR curve are exhibited in Fig. 5. As exhibited in Fig. 5, in this data set, the PSC is the best one among six methods, especially superior to SVM. The PSC has absolute advantages on the measures of CCRs, MWEs and BCCRs. The confusion matrix is exhibited in Table 4.

Table 4. The confusion matrix on Golub data set

| Method | Background | Classification | | CCR | Total CCR | 1-MWE | BCCR |
|---|---|---|---|---|---|---|---|
| | | True | False | | | | |
| DWD | TRUE | 426 | 24 | 0.946667 | 0.969907 | 0.964468 | 0.963857 |
| | FALSE | 15 | 831 | 0.98227 | | | |
| wDWD | TRUE | 426 | 24 | 0.946667 | 0.969907 | 0.964468 | 0.963857 |
| | FALSE | 15 | 831 | 0.98227 | | | |
| SVM | TRUE | 278 | 172 | 0.617778 | 0.800154 | 0.75747 | 0.728477 |
| | FALSE | 87 | 759 | 0.897163 | | | |
| PGLMC | TRUE | 422 | 28 | 0.937778 | 0.942901 | 0.941702 | 0.941673 |
| | FALSE | 46 | 800 | 0.945626 | | | |
| NPDMD | TRUE | 410 | 40 | 0.911111 | 0.957562 | 0.94669 | 0.944297 |
| | FALSE | 15 | 831 | 0.98227 | | | |
| PSC | TRUE | 431 | 19 | 0.957778 | **0.971451** | **0.968251** | **0.968038** |
| | FALSE | 18 | 828 | 0.978723 | | | |

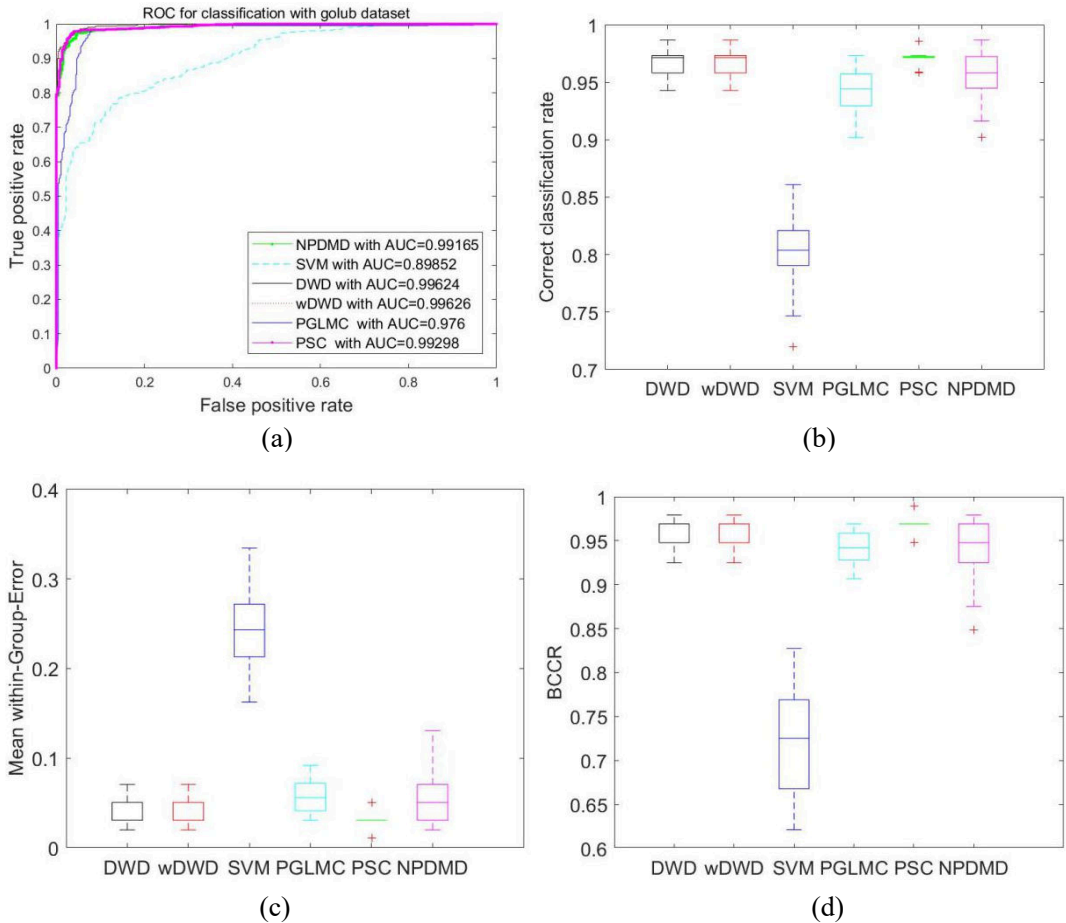

(a)  (b)  (c)  (d)

Figure 5. Comparison between six methods on Golub data for experiment 4. (a) ROC curves and AUC. (b) The boxplots of CCRs. (c) The boxplot for the mean within-group error. (d) The boxplot for BCCR.

### 4.3.4 Experiment 5: Gordon data set

In the Gordon data set [40], there are 181 tissue specimens, which include 31 malignant pleural mesotheliomas and 150 adenocarcinomas with imbalance factor $m \approx 4.84$. For each specimen, 12533 genes are expressed with U95A oligonucleotide probe arrays (Affymetrix, Santa Clara, CA).

The CCRs, MWEs and ROC curve are exhibited in Fig. 6. As exhibited in Fig. 6, in this data set, the PSC is the best one among six methods, especially superior to SVM.

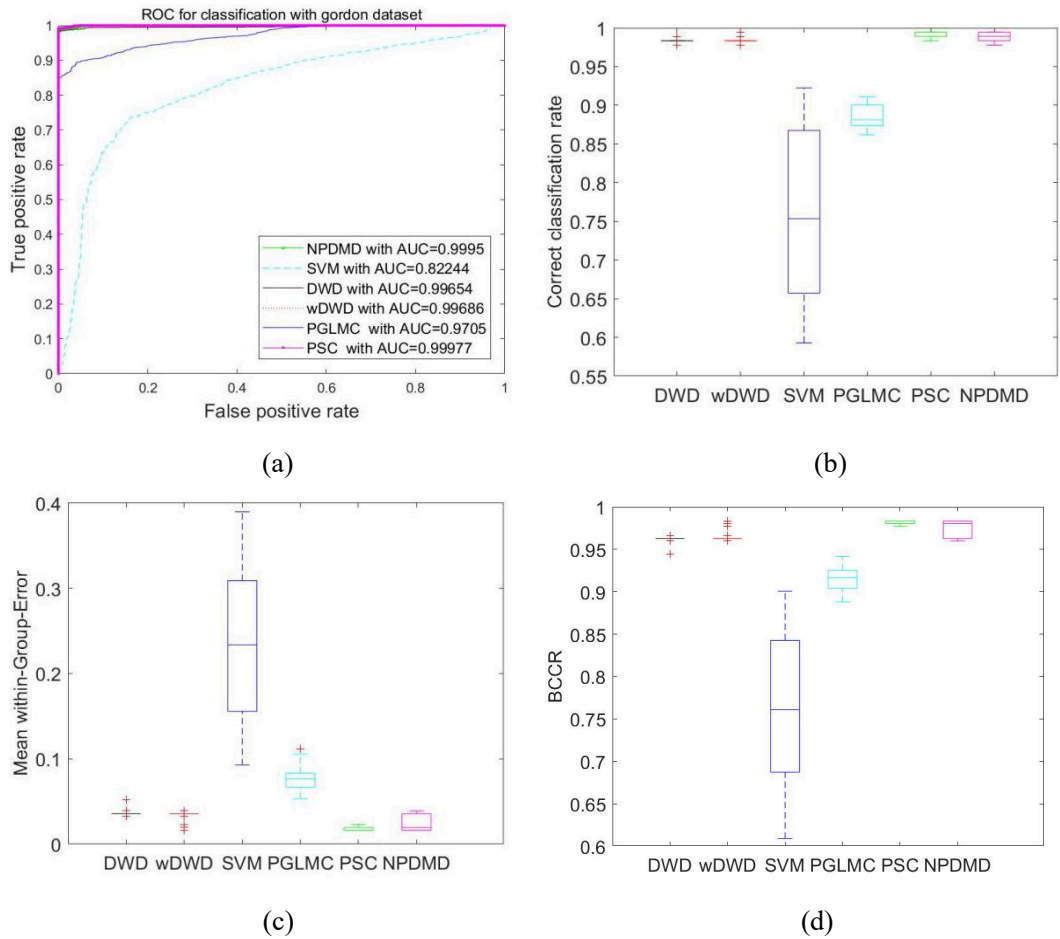

Figure 6. Comparison between six methods on Gordon data for experiment 5. (a) ROC curves and AUC. (b) The boxplots of CCRs. (c) The boxplot for the mean within-group error. (d) The boxplot for BCCR.

The NPDMD is suboptimal. The confusion matrix is shown in Table 5.

Table 5. The confusion matrix on Gordon data set

| Method | Background | Classification | | CCR | Total CCR | 1-MWE | BCCR |
|---|---|---|---|---|---|---|---|
| | | True | False | | | | |
| DWD | TRUE | 521 | 37 | 0.933692 | 0.983425 | 0.963698 | 0.961964 |
| | FALSE | 17 | 2683 | 0.993704 | | | |
| wDWD | TRUE | 525 | 33 | 0.94086 | 0.984039 | 0.966912 | 0.9656 |
| | FALSE | 19 | 2681 | 0.992963 | | | |
| SVM | TRUE | 437 | 121 | 0.783154 | 0.761203 | 0.76991 | 0.76964 |
| | FALSE | 657 | 2043 | 0.756667 | | | |
| PGLMC | TRUE | 545 | 13 | 0.976703 | 0.884592 | 0.921129 | 0.915457 |
| | FALSE | 363 | 2337 | 0.865556 | | | |
| NPDMD | TRUE | 532 | 26 | 0.953405 | 0.988643 | 0.974665 | 0.973785 |
| | FALSE | 11 | 2689 | 0.995926 | | | |
| PSC | TRUE | 540 | 18 | 0.967742 | **0.99202** | **0.982389** | **0.981968** |
| | FALSE | 8 | 2692 | 0.997037 | | | |

4.3.5 Experiment 6: Tian data set

In the Tian data set[42], there are 173 specimens of the purified plasma cells, which include 137 specimens with focal bone lesions and 36 without focal bone lesions with imbalance factor $m \approx 3.81$. For each specimen, 12625 genes are expressed with Affymetrix U95Av2 microarrays.

As exhibited in Fig. 7, in this data set, the PSC obtains the best performances on CCR and MWE, but suboptimal on BCCR. However, the PGLMC is optimal on BCCR. The confusion matrix is shown in Table 6. According to Table 6, the BCCR is a more reasonable measure than the CCR and MWE for IHDLSS data set.

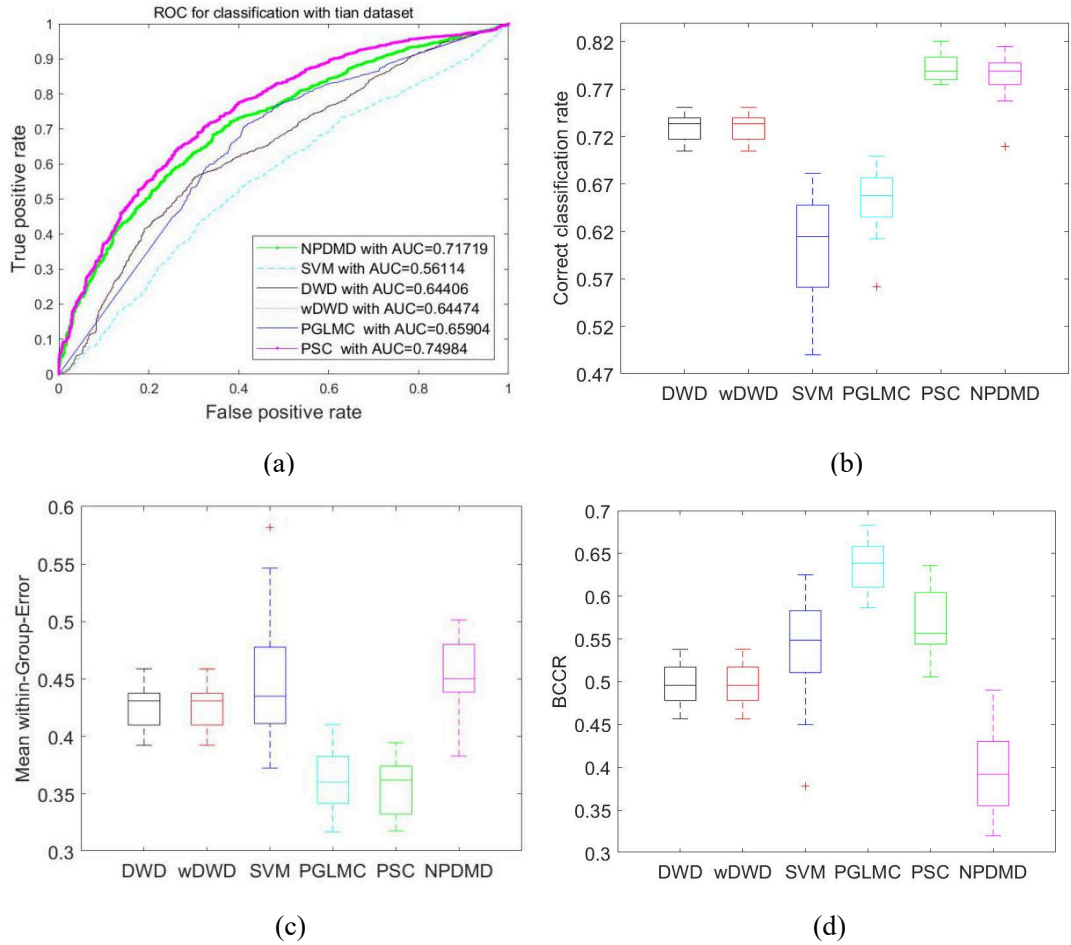

Figure 7. Comparison between six methods on Tian data for experiment 6. (a) ROC curves and AUC. (b) The boxplots of CCRs. (c) The boxplot for the mean within-group error. (d) The boxplot for BCCR.

Table 6. The confusion matrix on Tian data set

| Method | Background | Classification | | CCR | Total CCR | 1-MWE | BCCR |
|---|---|---|---|---|---|---|---|
| | | True | False | | | | |
| DWD | TRUE | 197 | 451 | 0.304012 | 0.73025 | 0.573134 | 0.495846 |
| | FALSE | 389 | 2077 | 0.842255 | | | |
| wDWD | TRUE | 197 | 451 | 0.304012 | 0.73025 | 0.573134 | 0.495846 |
| | FALSE | 389 | 2077 | 0.842255 | | | |
| SVM | TRUE | 304 | 344 | 0.469136 | 0.599872 | 0.551681 | 0.544214 |
| | FALSE | 902 | 1564 | 0.634225 | | | |
| PGLMC | TRUE | 391 | 257 | 0.603395 | 0.656712 | 0.637058 | **0.635616** |
| | FALSE | 812 | 1654 | 0.670722 | | | |
| NPDMD | TRUE | 91 | 557 | 0.140432 | 0.784522 | 0.547102 | 0.393025 |
| | FALSE | 114 | 2352 | 0.953771 | | | |
| PSC | TRUE | 253 | 395 | 0.390432 | **0.79255** | **0.644324** | 0.566388 |
| | FALSE | 251 | 2215 | 0.898216 | | | |

4.3.6 Experiment 7: Yeoh data set

The Yeoh data set [43] involves the diagnostic bone marrow samples from 248 pediatric acute lymphoblastic leukemia (ALL) patients who were determined to have one and only one of the six known pediatric ALL prognostic subtypes, which include T-cell lineage ALL (T-ALL), E2A-PBX1, TEL-AML1, MLL rearrangements, BCR-ABL, and hyperdiploid karyotypes with more than 50 chromosomes (HK50). The 248 patients include 43 T-ALL, 27 E2A-PBX1, 79 TEL-AML1, 15 BCR-ABL, 20 MLL, and 64 HK50 patients.

To construct a reasonable imbalanced set, the positive class is comprised of 27 E2A-PBX1, and the negative class is made up of the rest of the specimens (221 specimens). Therefore, the imbalance factor m≈8.19. The CCRs, ROC BCCR and MWEs are exhibited in Fig. 8.

As exhibited in Fig. 8, in the Yeoh data set, the PSC is the best one among six methods on all measurements of CCR, MWE and BCCR. The confusion matrix is exhibited in Table 7.

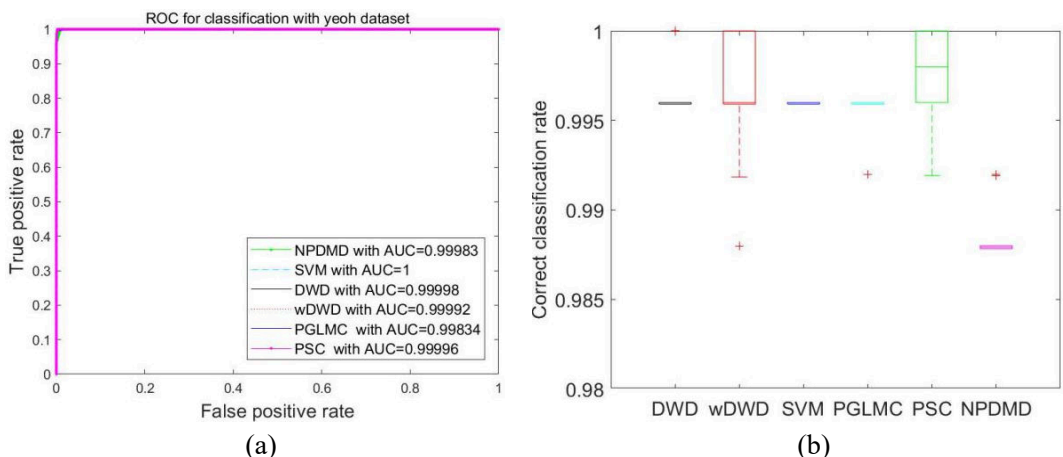

(a)           (b)

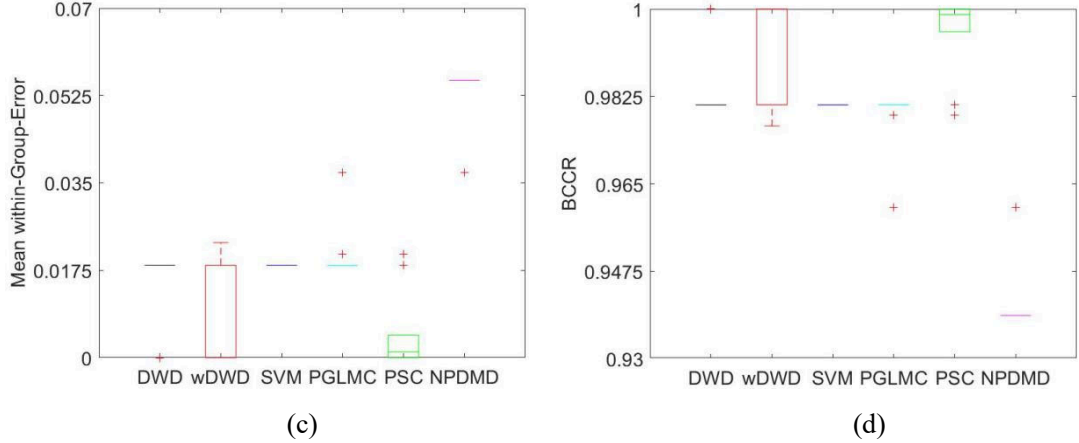

|     | (c) | (d) |

Figure 8. Comparison between six methods on Yeoh data for experiment 7. (a) ROC curves and AUC. (b) The boxplots of CCRs. (c) The boxplot for the mean within-group error. (d) The boxplot for BCCR.

Table 7. The confusion matrix on Yeoh data set

| Method | Background | Classification | | CCR | Total CCR | 1-MWE | BCCR |
|---|---|---|---|---|---|---|---|
|  |  | True | False |  |  |  |  |
| DWD | TRUE | 3978 | 0 | 1 | 0.996192 | 0.98251 | 0.981909 |
|  | FALSE | 17 | 469 | 0.965021 |  |  |  |
| wDWD | TRUE | 3973 | 5 | 0.998743 | 0.995968 | 0.985997 | 0.985677 |
|  | FALSE | 13 | 473 | 0.973251 |  |  |  |
| SVM | TRUE | 3978 | 0 | 1 | 0.995968 | 0.981481 | 0.980809 |
|  | FALSE | 18 | 468 | 0.962963 |  |  |  |
| PGLMC | TRUE | 3977 | 1 | 0.999749 | 0.99552 | 0.980327 | 0.979588 |
|  | FALSE | 19 | 467 | 0.960905 |  |  |  |
| NPDMD | TRUE | 3978 | 0 | 1 | 0.988575 | 0.947531 | 0.942328 |
|  | FALSE | 51 | 435 | 0.895062 |  |  |  |
| PSC | TRUE | 3971 | 7 | 0.99824 | **0.997536** | **0.995005** | **0.994984** |
|  | FALSE | 4 | 482 | 0.99177 |  |  |  |

### 4.3.7 Experiment 8: Burczynski data set

In Burczynski data set [44], transcriptional profiles in peripheral blood mononuclear cells were assessed, involving 42 healthy individuals, 59 Crohn's disease (CD) patients, and 26 ulcerative colitis (UC) patients for 22,283 gene expression levels.

To construct a reasonable imbalanced set, the positive class composes of 42 healthy
individuals and the negative class composes of 26 ulcerative colitis patients.

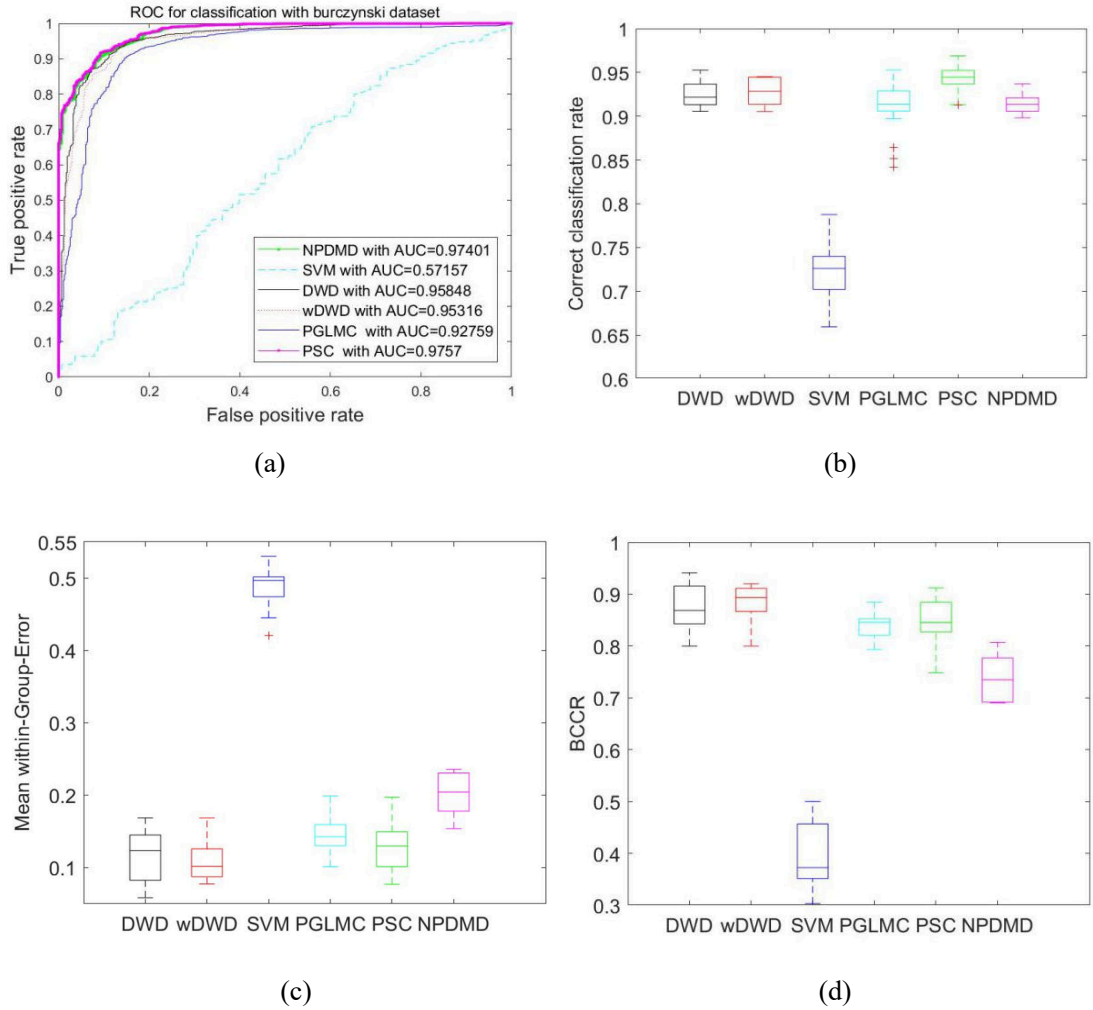

Figure 9. Comparison between six methods on Burczynski data. (a) ROC curves and AUC. (b) The boxplots of CCRs. (c) The boxplot for the mean within-group error. (d) The boxplot for BCCR.

individuals and 59 CD, and the negative class is made up of 26 UC. Therefore, the imbalance factor $m \approx 3.88$. The CCRs, ROC, BCCR and MWEs are exhibited in Fig. 9.

As exhibited in Fig. 9, in the Burczynski data set, the PSC obtains the best CCR. But the MWE and BCCR of the PSC are less than those of wDWD. We give the confusion matrix in Table 8.

Table 8. The confusion matrix on Burczynski data set

| Method | Background | Classification | | CCR | Total CCR | 1-MWE | BCCR |
|---|---|---|---|---|---|---|---|
| | | True | False | | | | |
| DWD | TRUE | 379 | 89 | 0.809829 | 0.925634 | 0.882637 | 0.873329 |
| | FALSE | 81 | 1737 | 0.955446 | | | |
| wDWD | TRUE | 389 | 79 | 0.831197 | 0.926509 | **0.891121** | **0.884744** |
| | FALSE | 89 | 1729 | 0.951045 | | | |
| SVM | TRUE | 71 | 397 | 0.151709 | 0.725722 | 0.512598 | 0.395049 |
| | FALSE | 230 | 1588 | 0.873487 | | | |
| PGLMC | TRUE | 358 | 110 | 0.764957 | 0.909011 | 0.855526 | 0.841605 |
| | FALSE | 98 | 1720 | 0.946095 | | | |
| NPDMD | TRUE | 281 | 187 | 0.600427 | 0.915573 | 0.798564 | 0.738262 |
| | FALSE | 6 | 1812 | 0.9967 | | | |
| PSC | TRUE | 353 | 115 | 0.754274 | **0.94007** | 0.871086 | 0.847635 |
| | FALSE | 22 | 1796 | 0.987899 | | | |

4.3.8 Experiment 9: Nakayama data set

In the Nakayama data set [45], the total RNA was isolated using TRIzol (Invitrogen, Carlsbad, CA, USA). A total of 105 samples from 10 types of soft tissue tumors were analyzed with a GeneChip Human Genome U133A array (Affymetrix, Santa Clara, CA, USA) containing 22283 probe sets.

To construct a reasonable imbalanced set, the positive class composes of the fifth type of tumors (21 specimens), and the negative class is made up of the rest of the specimens (84 specimens). Therefore, the imbalance factor $m=4$. The CCRs, ROC, BCCR and MWEs are exhibited in Fig. 10. As exhibited in Fig. 10, in the Nakayama data set, the PSC is also the best one among six methods on all measurements of CCR, MWE and BCCR. We show the confusion matrix in Table 9.

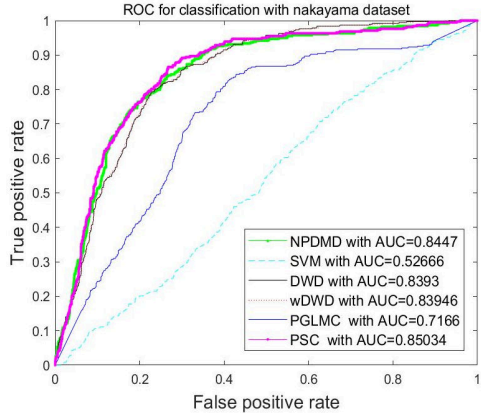 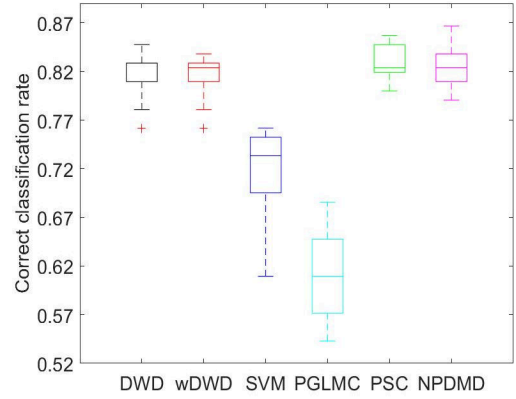

(a) (b)

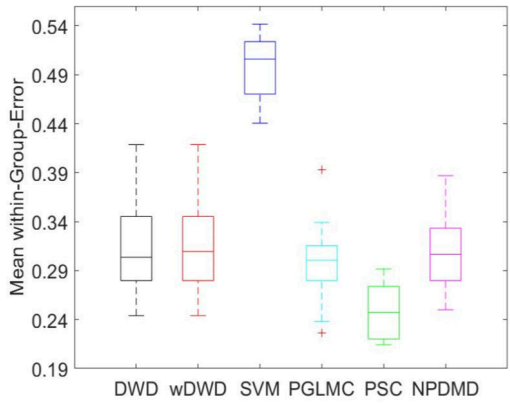 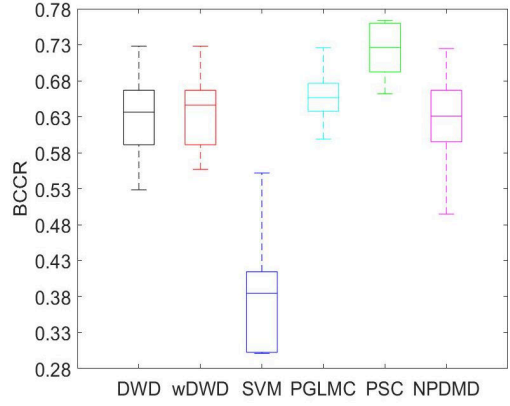

(c) (d)

Figure 10. Comparison between six methods on Nakayama data for experiment 9. (a) ROC curves and AUC. (b) The boxplots of CCRs. (c) The boxplot for the mean within-group error. (d) The boxplot for BCCR.

Table 9. The confusion matrix on Nakayama data set

| Method | Background | Classification | | CCR | Total CCR | 1-MWE | BCCR |
|---|---|---|---|---|---|---|---|
| | | True | False | | | | |
| DWD | TRUE | 1363 | 183 | 0.88163 | 0.80383 | 0.686929 | 0.636773 |
| | FALSE | 196 | 190 | 0.492228 | | | |
| wDWD | TRUE | 1357 | 189 | 0.877749 | 0.80176 | 0.687579 | 0.639603 |
| | FALSE | 194 | 192 | 0.497409 | | | |
| SVM | TRUE | 1308 | 204 | 0.865079 | 0.719048 | 0.5 | 0.383003 |
| | FALSE | 327 | 51 | 0.134921 | | | |
| PGLMC | TRUE | 827 | 685 | 0.546958 | 0.608466 | 0.700728 | 0.668361 |
| | FALSE | 55 | 323 | 0.854497 | | | |
| NPDMD | TRUE | 1379 | 133 | 0.912037 | 0.82381 | 0.691468 | 0.627357 |
| | FALSE | 200 | 178 | 0.470899 | | | |
| PSC | TRUE | 1334 | 178 | 0.882275 | **0.830159** | **0.751984** | **0.726882** |
| | FALSE | 143 | 235 | 0.621693 | | | |

4.4 Discussion

Tables 10 summarize the detailed results on the above eight data sets to make a comprehensive analysis. (Table 10 is in the last page). In Table 10, the best results for six methods on each measurement are marked in red. As can be seen, the overall performance of PSC are superior or highly competitive to the other compared methods. Specifically, for CCR, PSC performs significantly better than DWD/wDWD/SVM/PGLMC/NPDMD on 7/7/8/7/6 over 8 data sets respectively, and gets the best accuracy on 6 data sets; for MWE, PSC performs significantly better than DWD/wDWD/SVM/PGLMC/ NPDMD on 6/6/8/8/8 over 8 data sets respectively, and achieves the best accuracy on 6 data sets; for BCCR, PSC performs significantly better than DWD/wDWD/SVM/PGLMC/ NPDMD on 6/6/8/7/8 over 8 data sets respectively, and obtains the best accuracy on 5 data sets. In addition, as can be seen, in comparing with other methods, the average accuracy and the win/tie/loss counts of PSC are always better or comparable, almost never worse than other methods. It is clear that PSC should be the best one or approximate on each real IHDLSS dataset.

## 5. Conclusion

For the IHDLSS problem, the existing methods (such as SVM, DWD, wDWD, PGLMC and NPDMD) are subject to certain drawback. In this paper, we proposed a new cost-sensitive linear binary classifier PSC for IHDLSS. With the analysis on the state of the art methods for IHDLSS, PSC tries to maximize the sum of between-class scatter matrix and within-class scatter matrix on the premise of class separability, and assigns different intercept values to majority and minority classes. Due to this structural design, PSC displays superior performance in IHDLSS.

The major advantages of this study were four-fold. First, it works well on IHDLSS. Second, it is self-adaptive to determine the intercept term $b_\pm$ for each class. Third, the implement of PSC is easy and holds low computational complexity owing to solve the similar Convex Quadratic Programming formulation as in SVM. Fourth, the inverse of high dimensional matrix can be solved in low dimensional space. The experiment results manifested the superiority of PSC compared to the state-of-art algorithms in IHDLSS. Actually, these exhibit that it is a very promising linear binary classification, which is with great potential in many applications regardless of IHDLSS. Note that although we only consider linear binary classifier for simplicity in this paper, PSC can be extended to a kernel approach by a nonlinear mapping or work for multiclass even multi-label learning as SVM.

## ACKNOWLEDGMENT

The authors would like to thank (1) Prof. J.S. Marron in Department of biostatistics, University of North Carolina at Chapel Hill; (2) Dr. Boxiang Wang in Department of Statistics and Actuarial Science at University of Iowa; (3) Prof. Xingye Qiao in Department of Mathematical Sciences, Binghamton University, for their kind help on discussion about the methods based on Distance weighting.

Table 10. Comparison on 8 data sets. win/tie/loss counts for PSC are summarized in the last row.

| Datasets | Data | | CCR of Methods | | | | | | 1-MWE | | | | | | BCCR | | | | | |
|---|---|---|---|---|---|---|---|---|---|---|---|---|---|---|---|---|---|---|---|---|
| | Dim | Classes | DWD | wDWD | SVM | PGLMC | NPDMD | PSC | DWD | wDWD | SVM | PGLMC | NPDMD | PSC | DWD | wDWD | SVM | PGLMC | NPDMD | PSC |
| Alon | 2000 | 2 | 0.789 | 0.784 | 0.545 | **0.813** | 0.812 | 0.807 | 0.771 | 0.768 | 0.504 | 0.784 | 0.789 | **0.798** | 0.766 | 0.763 | 0.485 | 0.770 | 0.780 | **0.797** |
| shipp | 7129 | 2 | **0.962** | **0.962** | 0.879 | 0.776 | 0.961 | 0.957 | **0.958** | **0.958** | 0.859 | 0.850 | 0.935 | 0.954 | **0.958** | **0.958** | 0.856 | 0.814 | 0.930 | 0.953 |
| Golub | 7129 | 2 | 0.970 | 0.970 | 0.800 | 0.943 | 0.958 | **0.971** | 0.964 | 0.964 | 0.757 | 0.942 | 0.947 | **0.968** | 0.964 | 0.964 | 0.728 | 0.942 | 0.944 | **0.968** |
| Gordon | 12533 | 2 | 0.983 | 0.984 | 0.761 | 0.885 | 0.989 | **0.992** | 0.964 | 0.967 | 0.770 | 0.921 | 0.975 | **0.982** | 0.962 | 0.966 | 0.770 | 0.915 | 0.974 | **0.982** |
| tian | 12625 | 2 | 0.730 | 0.730 | 0.600 | 0.657 | 0.785 | **0.793** | 0.573 | 0.573 | 0.552 | 0.637 | 0.547 | **0.644** | 0.496 | 0.496 | 0.544 | **0.636** | 0.393 | 0.566 |
| yeoh | 12625 | 6 | 0.996 | 0.996 | 0.996 | 0.996 | 0.989 | **0.998** | 0.983 | 0.986 | 0.981 | 0.980 | 0.948 | **0.995** | 0.982 | 0.986 | 0.981 | 0.980 | 0.942 | **0.995** |
| Burczynski | 22283 | 3 | 0.926 | 0.927 | 0.726 | 0.909 | 0.916 | **0.940** | 0.883 | **0.891** | 0.513 | 0.856 | 0.799 | 0.871 | 0.873 | **0.885** | 0.395 | 0.842 | 0.738 | 0.848 |
| nakayama | 22283 | 10 | 0.804 | 0.802 | 0.719 | 0.608 | 0.824 | **0.830** | 0.687 | 0.688 | 0.500 | 0.701 | 0.691 | **0.752** | 0.637 | 0.640 | 0.383 | 0.668 | 0.627 | **0.727** |
| | | | | | | | | | | | | | | | | | | | | |
| Average accuracy | | | 0.895 | 0.894 | 0.753 | 0.823 | 0.904 | **0.911** | 0.848 | 0.849 | 0.679 | 0.834 | 0.829 | **0.871** | 0.830 | 0.832 | 0.643 | 0.821 | 0.791 | **0.855** |
| PSC：W/T/L | | | 7/0/1 | 7/0/1 | 8/0/0 | 7/0/1 | 6/0/2 | | 6/0/2 | 6/0/2 | 8/0/0 | 8/0/0 | 8/0/0 | | 6/0/2 | 6/0/2 | 8/0/0 | 7/0/1 | 8/0/0 | |